# SELF-SUPERVISED SPATIAL AND ZERO-SHOT ANGULAR SUPER-RESOLUTION BY SPATIAL-ANGULAR IMPLICIT REPRESENTATION FOR ROTATING-VIEW SNR-EFFICIENT DIFFUSION MRI

*Yinzhe Wu*[1,2]*, Hongyu Rui*[1]*, Fanwen Wang*[1,2]*, Jiahao Huang*[1,2]*, Zi Wang*[1]*, Guang Yang*[1,2]

[1]Department of Bioengineering and I-X, Imperial College London, London, UK
[2]Cardiovascular Research Centre, Royal Brompton Hospital, London, UK
{yinzhe.wu18, g.yang}@imperial.ac.uk

## ABSTRACT

Rotating-view thick-slice acquisition is highly SNR-efficient for mesoscale diffusion MRI (dMRI) but requires numerous rotating views to satisfy Nyquist sampling, resulting in long scan time. We propose a self-supervised Spatial-Angular Implicit Neural Representation (SA-INR) that reconstructs high-resolution dMRI from a single view per diffusion direction, representing a massive acceleration. Our model, an MLP conditioned on a $b = 0$ structural prior and the b-direction via FiLM, is trained end-to-end on the anisotropic input. The framework not only accurately reconstructs the trained b-directions (spatial SR) but also learns a continuous q-space representation, enabling high-fidelity *"zero-shot"* synthesis of unseen b-directions (angular SR). On simulated data, our method achieved high fidelity for both trained (34.82 dB) and unseen (33.08 dB) directions. Most importantly, the synthesized angular data also improved the quantitative accuracy of downstream DTI model fitting. Our SA-INR framework breaks the classical sampling limits, paving the way for fast, quantitative high-resolution dMRI.

## 1. INTRODUCTION

The evolution of *in-vivo* diffusion MRI (dMRI) has progressively provided valuable insights into the structural connectivity and tissue microstructure of the human brain. Recently, high-resolution (HR) dMRI techniques have advanced the field by enabling the investigation of fine-scale features such as intracortical layers and short-range U-fibers. When combined with advanced diffusion modeling, high b-value acquisitions allow more comprehensive and specific characterization of tissue microstructure, offering enhanced potential to reveal detailed information about the brain in both health and disease. However, substantial technical challenges persist as spatial resolution and b-values increase, since both lead to a diminished signal-to-noise ratio (SNR) owing to smaller voxel volumes and stronger diffusion-encoding attenuation.

To overcome these limitations, several acquisition strategies have sought to improve SNR efficiency without compromising resolution. Multi-shot 2D and 3D acquisitions such as multi-slab EPI [1] and 3D MERMAID [2] exploit volumetric encoding and optimized recovery to boost SNR per unit time. Likewise, the Romer-EPTI [3] framework introduced a *rotating-view thick-slice* acquisition scheme that achieves mesoscale resolution by acquiring multiple thick-slice volumes at different orientations and reconstructing a high-isotropic image through motion-aware super-resolution. These rotating-view approaches are particularly attractive because they preserve high in-plane resolution while recovering isotropic detail through thick-slice averaging, effectively trading anisotropy for SNR efficiency.

Despite their success, these methods remain hampered by the large number of slice orientations required to satisfy the Nyquist criterion for isotropic reconstruction. For instance, Romer-EPTI must acquire at least $\mathrm{N_{min}} = (\pi/2)\mathrm{t_s}$ rotating views for each diffusion-encoding direction, where $t_s$ is the slice-thickness factor (i.e., the ratio between the acquired thick slice and the target isotropic voxel size) [3], [4]. With typical parameters ($\mathrm{t_s} = 8$), this necessitates $N \geq 12$ orientations per diffusion direction and total scan times exceeding an hour, prohibitive for routine clinical use. Thus, while rotating-view thick-slice imaging achieves the highest SNR efficiency per acquisition, the overall throughput remains constrained by extensive directional repetitions.

In this work, we introduce a conceptually different strategy that exploits the ***implicit complementarity*** between $q$-space diffusion-encoding and $x$-space slice-orientation domains. Instead of acquiring multiple rotating-view thick slices for each diffusion direction, we acquire only one thick-slice orientation per diffusion direction, with each direction using a distinct slice-rotation vector. Because different diffusion gradients ($q$-vectors) already sample the signal along independent orientations in $q$-space, their associated thick-slice orientations collectively provide complementary spatial sampling across the 3D volume. By coupling these anisotropic acquisitions through a unified ***spatial-angular implicit representation*** trained in a self-supervised manner, we can reconstruct an isotropic, high-resolution diffusion signal field across both spatial ($x$) and angular ($q$) domains. This design effectively reduces the required number of thick-slice views by a factor of $N_{min}$ or more, achieving substantial acceleration while maintaining SNR efficiency.

Our framework fuses information from all diffusion directions through a feature-wise linear modulation (FiLM) conditioned implicit neural representation (INR) that maps continuous spatial coordinates ($\mathbf{x}$) and diffusion directions ($\mathbf{g}$) to their corresponding signal intensities. The high-SNR b = 0 reference image serves as a structural prior, conditioning the INR to preserve anatomical detail. Beyond accelerating rotating-view thick-slice dMRI, the **continuous *x-q* representation** inherently enables *zero-shot angular super-resolution*, allowing inference of unseen diffusion directions without additional acquisitions. Together, these innovations provide a new pathway for SNR-efficient, high-resolution, and directionally continuous diffusion MRI within clinically feasible scan times.

## 2. METHOD

Our proposed method reconstructs a HR isotropic dMRI volume from a set of highly undersampled, anisotropic, rotating-view thick-slice acquisitions. We achieve this by formulating the reconstruction as a self-supervised, multi-contrast super-resolution problem, solved using a spatial-angular implicit neural representation (SA-INR) (**Fig.1**).

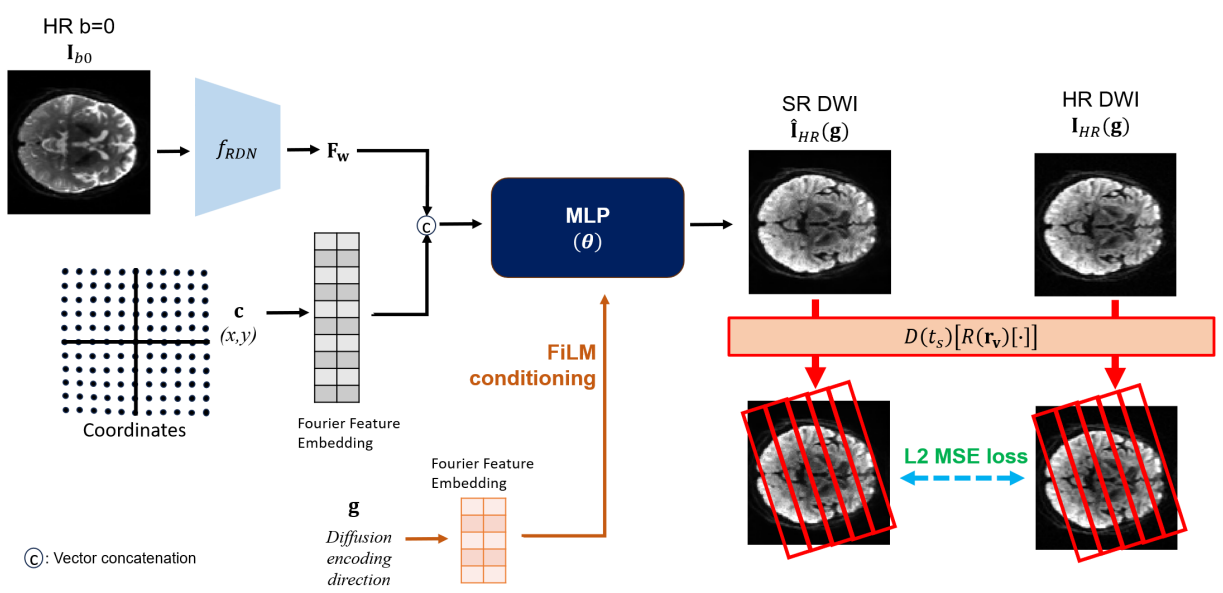

**Fig. 1** Overview of the Self-Supervised Spatial-Angular Implicit Neural Representation Framework. Reproduced by kind permission of UK Biobank ©

### 2.1. Problem Formulation and Forward Model

Our goal is to estimate the set of HR isotropic DWIs, $I_{HR}(\mathbf{g})$, for all diffusion-encoding b-directions $\mathbf{g} \in \mathcal{G}$. The acquired data consists of a set of low-resolution (LR), high-SNR, anisotropic thick-slice images, $I_{LR}(\mathbf{g}, \mathbf{r_v})$, where $\mathbf{r_v} \in \mathcal{R}$ is the rotation-view direction (i.e., the axial vector of the anisotropic slice) and $t_s$ is the slice thickness factor.

The forward model $M$ that maps the HR image to a single LR acquisition can be expressed as:

$$I_{LR}(\mathbf{g}, \mathbf{r_v}) = M\big(I_{HR}(\mathbf{g})\big) + \mathbf{n} \quad (1)$$

where $M(\cdot) = D(t_s) \circ R(\mathbf{r_v})(\cdot)$ is a composite operator. $R(\mathbf{r_v})$ is a 3D rotation of the volume by the view-angle $\mathbf{r_v}$, and $D(t_s)$ is the thickness factor $t_s$ dependent anisotropic downsampling operator.

Classical super-resolution methods require a sufficient number of rotating views $N$ to satisfy the Nyquist sampling criterion, $N \geq (\pi/2)t_s$, to solve this ill-posed problem for each $\mathbf{g}$ independently. To dramatically accelerate this, we propose to solve for all $I_{HR}(\mathbf{g})$ jointly from a highly undersampled set of views (e.g., $N = 1$), far below the Nyquist limit.

### 2.2. Spatial-Angular Implicit Neural Representation (SA-INR)

We model the entire dMRI volume ($x$- & $q$-space) as a continuous function $f_\theta$, approximated by a coordinate-based multi-layer perceptron (MLP) with parameters θ. For a given 2D slice, this function maps a spatial coordinate $\mathbf{c} = (x, y)$ and a b-direction $\mathbf{g}$ to the corresponding HR DWI intensity:

$$\hat{I}_{HR}(\mathbf{c}, \mathbf{g}) = f_\theta(\mathbf{c}, \mathbf{F_w}, \mathbf{g}) \quad (2)$$

The network is conditioned on both a structural prior $F_w$ and the diffusion direction $\mathbf{g}$.

1. ***Spatial Encoding and Structural Prior:*** The 2D spatial coordinates $\mathbf{c}$ are first mapped to a high-dimensional space using a Fourier Feature embedding, $\gamma(\boldsymbol{c})$, to enable the network to learn high-frequency details. To provide a high-fidelity anatomical scaffold, we exploit the high-SNR $b = 0$ image $I_{b=0}$. We pass $I_{b=0}$ through a Residual Dense Network (RDN) to extract rich spatial features $\mathbf{F_w} = f_{RDN}(I_{b=0})$ [5]. At each coordinate $\mathbf{c}$, the corresponding feature vector $\boldsymbol{F_w}(\boldsymbol{c})$ is sampled and concatenated with the spatial embedding $\gamma(\boldsymbol{c})$ to form the primary spatial input to the MLP.

2. ***Angular Conditioning via FiLM:*** To make the network's output specific to the diffusion contrast, we condition it on the b-direction vector $\mathbf{g}$. The vector $\mathbf{g}$ is passed through its own Fourier Feature embedding, $\gamma(\mathbf{g})$. This embedding is then used to predict scaling (β) and shifting (α) parameters via small MLPs. These parameters modulate the intermediate activations $a_i$ of the main MLP at layer $i$ using Feature-wise Linear Modulation (FiLM) [6]:

$$\mathrm{FiLM}(a_i) = \alpha\big(\gamma(\mathbf{g})\big) \odot a_i + \beta\big(\gamma(\mathbf{g})\big) \quad (3)$$

This allows the b-direction $g$ to adaptively control the network's computation, effectively selecting the correct diffusion encoding contrast for the spatial reconstruction.

### 2.3. Self-Supervised Training

The network $f_\theta$ is trained in a self-supervised manner, requiring no HR ground truth. For each 2D slice, a separate INR is optimized. During training, we sample the full coordinate space $\mathbf{c} \in \mathbf{\Omega}$ and a batch of a b-direction $\mathbf{g_{train}}$ from the training set. The network predicts the HR patch $\hat{I}_{HR}(\mathbf{g_{train}})$. We then apply the known, non-trainable forward model $M$ (**Eq. 1**) to this patch to generate a predicted LR patch, $\hat{I}_{LR}$. The loss $\mathcal{L}$ is the L2 (MSE) loss between this $\hat{I}_{LR}$ and the corresponding patch from the actual acquired LR image $I_{LR}(\mathbf{g_{train}}, \mathbf{r_v})$:

$$\mathcal{L} = \mathbb{E}_{\mathbf{g},\mathbf{r_v}}\left[\sum_{\mathbf{c}\in\mathbf{\Omega}} \left\| M\big(f_\theta(\mathbf{c}, \mathbf{F_w}, \mathbf{g})\big) - I_{LR}(\mathbf{g}, \mathbf{r_v})(\mathbf{c}) \right\|_2^2\right] \quad (4)$$

The network parameters θ are optimized by minimizing this loss over all training b-directions and acquired views.

### 2.4. Zero-Shot Angular Super-Resolution

A key capability of our SA-INR is that $f_\theta$ learns a continuous representation of $q$-space. After training is complete, we can synthesize a high-resolution DWI for any unseen b-direction $\mathbf{g_{unseen}}$ simply by querying the network:

$$\hat{I}_{HR}(\mathbf{c}, \mathbf{g}_{\mathbf{unseen}}) = f_{\theta}(\mathbf{c}, \mathbf{F}_{\mathbf{w}}, \mathbf{g}_{\mathbf{unseen}}) \quad (5)$$

This inference is "zero-shot" as the network has never been trained on $\mathbf{g}_{\mathbf{unseen}}$.

### 2.5. Experimental Setup

**a) Data:** We used 2D axial slices from 8 participants from the UK Biobank. The original 2 mm isotropic data served as our HR ground truth (GT) and included an averaged $b = 0$ image and 50 $b = 1000$ s/mm$^2$ directions.

**b) Simulated Acquisition:** The 2 mm GT data was anisotropically downsampled by $t_s = 4$ (simulating 2×2×8 mm voxels). We simulated a 7× accelerated acquisition, where for the Nyquist limit of $N \geq (\pi/2)t_s \approx 7$ views we used only $\boldsymbol{N} = \mathbf{1}$ **rotating-view** per b-direction. The rotation vector $\mathbf{r_v}$ for each $\mathbf{g}$ was set as the 2D projection of $\mathbf{g}$ onto the $x$-$y$ plane.

**c) Training and Evaluation:** The 50 b-directions were split into 40 for training and 10 held-out directions for testing. We evaluated: **(i) Spatial SR:** Reconstruction quality (PSNR, SSIM, LPIPS) on the 40 trained directions. **(ii) Angular SR:** Zero-shot synthesis quality on the 10 held-out directions. **(iii) Downstream Fidelity:** We fit diffusion tensor imaging (DTI) models to *(1)* 40 trained SR DWIs and *(2)* 40 trained + 10 zero-shot SR DWIs, and compared the resulting FA and MD maps to a GT model fit using all 50 HR DWIs.

## 3. RESULT AND DISCUSSION

We evaluated our SA-INR framework by first training it on the 40 *"trained"* b-directions (self-supervised spatial SR) and then testing its ability to synthesize the 10 *"unseen"* held-out directions (zero-shot angular SR).

### 3.1. Self-Supervised Spatial Super-Resolution

Our model successfully reconstructed high-resolution, isotropic DWIs from the accelerated (*N=1*) anisotropic input. **Table I(A)** compares the quantitative metrics for the 40 trained b-directions against the bilinear baseline. Our method achieved a 5.65 dB gain in PSNR and a 7.5% improvement in SSIM, demonstrating a dramatic increase in reconstruction fidelity. Qualitative results **(Fig. 2(A))** confirm this. The reconstructed SR image is visually almost indistinguishable from the HR GT, with the absolute error map showing near-zero structural error. A representative line profile confirms the high-fidelity recovery of fine-scale intensity variations.

### 3.2. Zero-Shot Angular Super-Resolution

The network's ability to synthesize unseen b-directions was highly effective. As shown in **Table I(B)**, the zero-shot synthesized DWIs achieved a PSNR of 33.08 dB, significantly outperforming the 28.26 dB baseline. This demonstrates that the FiLM-conditioned network successfully learned a continuous and generalizable representation of $q$-space. Qualitatively **(Fig. 2(B))**, the synthesized unseen DWI shows high anatomical fidelity and contrast accuracy, with only a minor increase in error compared to the trained direction.

### 3.3. Downstream Quantitative Fidelity

**Table I.** Quantitative Results of Self-Supervised Spatial Super-Resolution. Evaluation metrics comparing the bilinear baseline (LR) against our SA-INR method (SR) on the 40 trained b-directions. *: $p$<0.01 from all other rows

| Mean(Std) | | DWI PSNR ↑ | DWI SSIM ↑ | DWI LPIPS ↓ | DWI/$S_0$ NMSE ↓ |
|---|---|---|---|---|---|
| (A) trained | LR | 29.17(1.37) | .8716(.0275) | .2367(.0350) | .0488(.0075) |
| | SR | ***34.82(0.93)** | ***.9466(.0053)** | ***.1224(.0188)** | ***.0435(.0053)** |
| | SR w/o b=0 | 34.65(0.99) | .9430(.0076) | .1268(.0174) | .0452(.0064) |
| (B) unseen | LR | 28.26(1.81) | .8440(.0461) | .2641(.0515) | .0488(.0077) |
| | SR | ***33.08(0.96)** | ***.9300(.0081)** | ***.1310(.0191)** | ***.0455(.0058)** |
| | SR w/o b=0 | 32.13(2.00) | .9216(.0163) | .1377(.0197) | .0481(.0077) |

**Table II**. Quantitative Results of Zero-Shot Angular Super-Resolution. Evaluation of the model's generalization capability on the 10 held-out (unseen) b-directions. *: $p$<0.01

| NMSE ↓ Mean(Std) | \|MD\| $\times 10^{-2}$ | \|FA\| $\times 10^{-2}$ | \|AD\| $\times 10^{-2}$ | \|RD\| $\times 10^{-2}$ | \|EV1\|×\|FA\| $\times 10^{-2}$ |
|---|---|---|---|---|---|
| **40 trained q-vecs** | **2.86(4.73)** | 4.29(0.748) | **4.35(5.74)** | **2.91(4.94)** | 9.69(1.76) |
| **40 trained + 10 unseen** | 2.95(4.81) | ***4.18(0.730)** | 4.48(5.51) | 2.97(5.05) | ***9.61(1.79)** |

**Fig. 2.** Visual comparison between the SA-INR Super-Resolution (1) reconstruction result (SR), (2) the High-Resolution Ground Truth (GT), and (3) the corresponding absolute error map, with their (4) line profile, for (A) self-supervised and (B) unseen ones. Reproduced by kind permission of UK Biobank ©

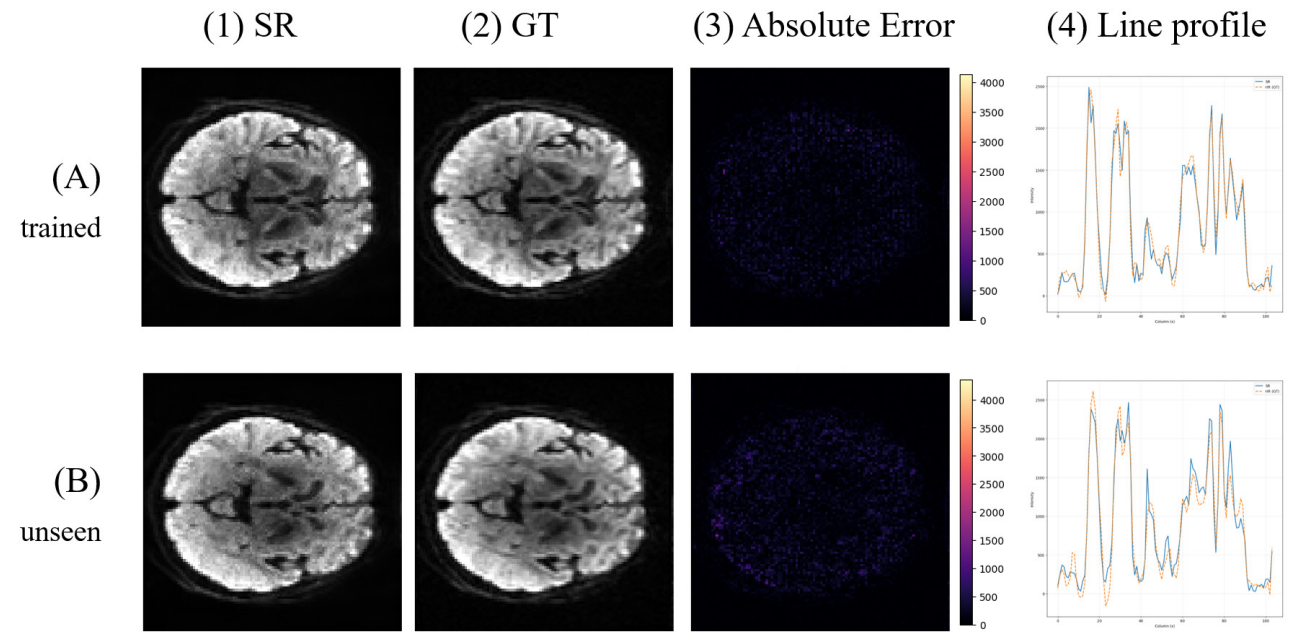

We assessed the practical utility of our method by fitting DTI models. **Fig. 3** shows that maps derived from our 40-direction SR are closely matching the 50-direction HR ground truth.

Most critically, we tested if the synthesized angular data was quantitatively meaningful. We compared the NMSE of maps from 40 SR directions against maps from 40 SR + 10 zero-shot directions. As shown in **Table II**, adding the 10 synthesized directions statistically improved ($p$<0.01) the NMSE for Fractional Anisotropy (FA) and the FA modulated principal eigenvector map (|EV1|×|FA|). Conversely, the metrics for Mean Diffusivity (MD), Axial Diffusivity (AD), and Radial Diffusivity (RD) showed statistically insignificant change, suggesting that the synthesis primarily refines the angular estimation, not the underlying average diffusivity.

### 3.4. Ablation Study

Finally, we validated the contribution of the $b = 0$ structural prior (**Table I**). A model trained without the RDN feature extractor performed worse than our full model, particularly on the more challenging zero-shot synthesis task. This confirms the structural prior is beneficial for guiding both spatial and angular reconstruction.

## 4. DISCUSSION

We have proposed a self-supervised, spatial-angular INR framework that successfully reconstructs high-resolution

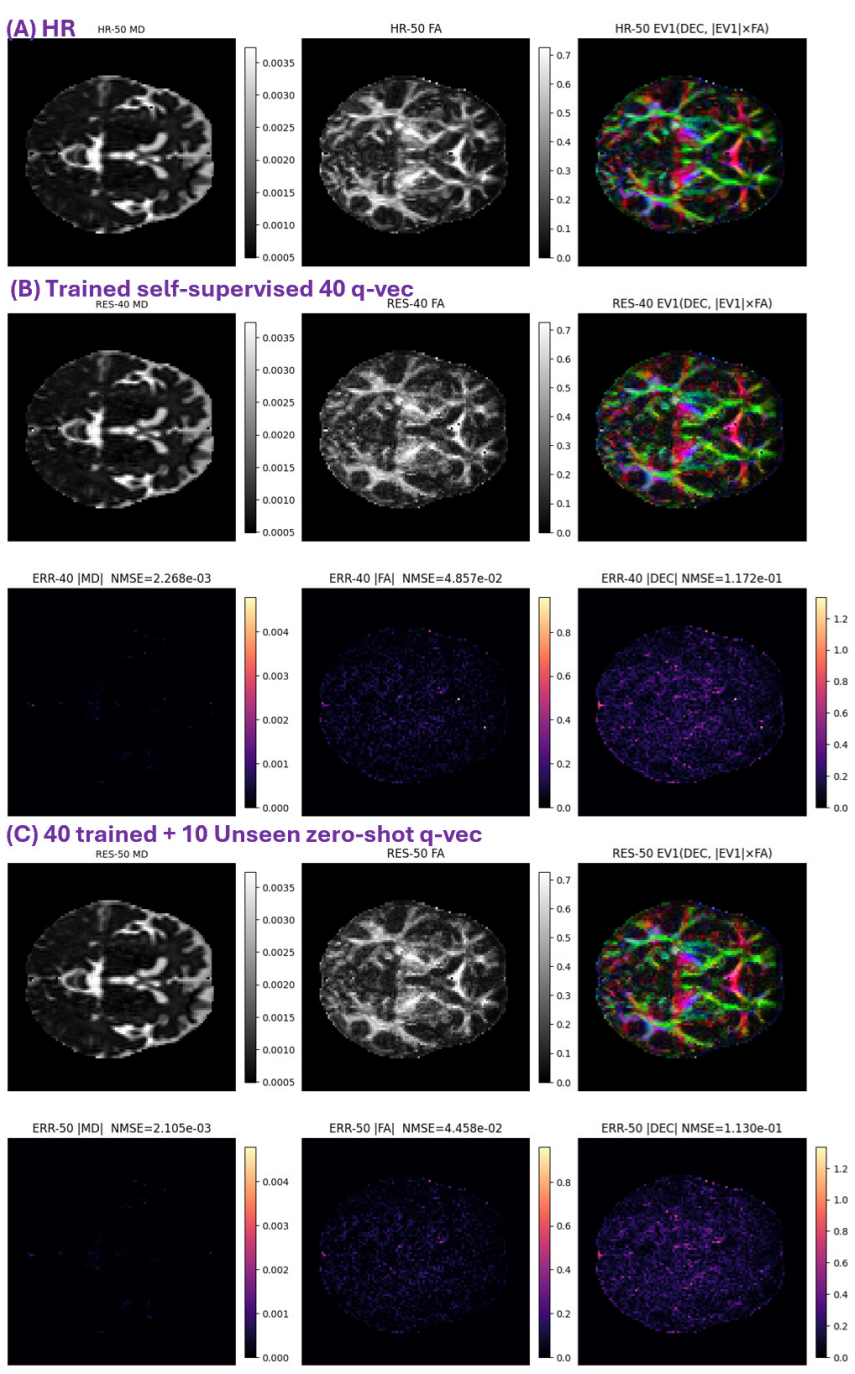

**Fig.3**. Downstream quantitative map fitting: Comparison of fitted Mean Diffusivity (MD), Fractional Anisotropy (FA) and FA modulated principal eigenvector (EV1) maps and their error maps, derived from (A) HR of fully sampled 50 b-directions, (B) 40 self-supervised b-direction DWI results (RES-40), and (C) 40 + 10 unseen b-direction DWI results (RES-50). Reproduced by kind permission of UK Biobank ©

dMRI from a single $N = 1$ rotating thick-slice view, representing 7× acceleration over the Nyquist sampling limit. The primary significance of this work is the demonstration that the classical sampling requirement for rotating-view super-resolution, $N \geq (\pi/2)t_s$, can be broken. The Romer-EPTI technique proved the SNR-efficiency of the rotating-view acquisition but required $N = 12$ views, leading to a ~78-minute scan [3]. Our SA-INR framework, by replacing the linear inverse problem with a continuous, non-linear function approximation, achieves high-fidelity spatial reconstruction from $N = 1$. This drastically reduces the data requirement and potentially opens a viable path to mesoscale dMRI in a clinically feasible scan time.

The most novel contribution is the zero-shot angular super-resolution capability. By explicitly conditioning the network on the b-vector **g** via FiLM, the network learns a continuous representation of $q$-space. The key finding (**Table II**) is that this synthesized data is not just visually plausible but quantitatively meaningful, significantly improving the stability and accuracy of downstream DTI model fitting. Specifically, the observed significant improvement in FA and EV1 fidelity confirms that the synthesized directions effectively constrain the angular uncertainty of the diffusion tensor. Crucially, the lack of significant change in MD, AD, and RD metrics supports the integrity of our method, as these average diffusivity measures are inherently less dependent on the angular density of the $q$-space sampling, verifying that the synthesis accurately preserves the fundamental magnitude of the signal attenuation while refining its directional component. This suggests a new acquisition paradigm: one could acquire a sparse set of b-directions and use the SA-INR to synthesize the dense $q$-space required for advanced microstructural models, achieving a "two-for-one" acceleration in both spatial and angular domains.

This study serves as a proof-of-concept and has limitations. The validation was a 2D simulation based on 2 mm isotropic data. Future work must validate this method on real, prospective, mesoscale *in-vivo* data (e.g., 500 µm data), where motion should also be modelled in the forward model as part of the composite operator $M(\cdot)$ **(Eq. 1)**, alike in [3]. Furthermore, the model must be extended from 2D-per-slice to a full 3D spatial model and expanded to handle multi-shell data by conditioning on both the b-vector **g** and the b-value.

## 5. CONCLUSION

We introduced a self-supervised, spatial-angular implicit neural representation (SA-INR) for rotating-view dMRI. Our framework achieves acceleration by reconstructing high-resolution DWIs from a single thick-slice rotating view per diffusion direction, breaking the classical Nyquist limit. It further provides a novel zero-shot angular super-resolution capability that synthetically densifies $q$-space and improves the accuracy of quantitative diffusion models. This work presents a new and powerful reconstruction paradigm for enabling ultra-fast, high-resolution, and quantitative dMRI.

## 6. ACKNOWLEDGMENTS

This research has been conducted using the UK Biobank Resource under Application Number 100203.

This study was supported in part by Imperial College London President's PhD Scholarship and in part by Imperial College London I-X. G. Yang was supported by UKRI Future Leaders Fellowship (MR/V023799/1, UKRI2738).